\pdfoutput=1

\documentclass[11pt]{article}

\usepackage[final]{acl}

\usepackage{times}
\usepackage{latexsym}

\usepackage[T1]{fontenc}

\usepackage[utf8]{inputenc}

\usepackage{microtype}

\usepackage{inconsolata}

\usepackage{graphicx}

\usepackage{subcaption}
\usepackage{mwe}

\usepackage{multirow}
\usepackage{microtype}

\usepackage{amsfonts}

\title{Disentangling Mathematical Reasoning in LLMs: A Methodological Investigation of Internal Mechanisms}

\newcommand{\affilsup}[1]{\rlap{\textsuperscript{\normalfont#1}}}

\author{
    Tanja Baeumel\affilsup{1, 2, 3}
    \qquad
    Josef van Genabith\affilsup{1, 2}
    \qquad
    Simon Ostermann\affilsup{1, 2, 3}
    \\
    $^1$German Research Center for Artificial Intelligence (DFKI) \\
    $^2$Saarland University \\
    $^3$Center for European Research in Trusted AI (CERTAIN) \\
    \texttt{tanja.baeumel@dfki.de} \\
}

\begin{document}
\maketitle
\begin{abstract}
Large language models (LLMs) have demonstrated impressive capabilities, yet their internal mechanisms for handling reasoning-intensive tasks remain underexplored. To advance the understanding of model-internal processing mechanisms, we present an investigation of how LLMs perform arithmetic operations by examining internal mechanisms during task execution. Using early decoding, we trace how next-token predictions are constructed across layers. Our experiments reveal that while the models recognize arithmetic tasks early, correct result generation occurs only in the final layers. Notably, models proficient in arithmetic exhibit a clear division of labor between attention and MLP modules, where attention propagates input information and MLP modules aggregate it. This division is absent in less proficient models. Furthermore, successful models appear to process more challenging arithmetic tasks functionally, suggesting reasoning capabilities beyond factual recall. 
\end{abstract}
\section{Introduction}

The increasingly impressive capabilities of large language models (LLMs) are generating a growing interest in the underlying mechanisms that facilitate their exceptional performance \cite{mosbach_insights_2024}. Recent work in interpretability 
has allowed the community to gradually build a better understanding of \textit{how} transformer-based models \cite{vaswani2017attention} retrieve and use factual information that is implicitly stored in their parameters \cite{geva_transformer_2022, meng_locating_2023, merullo_language_2024, hernandez_linearity_2024, nanda_emergent_2023, yu2023characterizing, todd_function_2024}. 
However, our understanding of the mechanisms employed by LLMs to solve non-factual and reasoning-intensive tasks is less well developed.

In this study, we investigate the mechanisms by which large language models (LLMs) tackle mathematical reasoning. We focus on a task that is both straightforward to evaluate and interpret: specifically, we investigate how LLMs perform basic arithmetic operations (e.g., \textit{Please calculate 143 + 81 = }) and analyze the differences in the internal mechanisms of models exhibiting different degrees of arithmetic proficiency.

We investigate the model-internal mechanisms via the interpretability method of \textit{early decoding} \cite{nostalgebraist, geva_transformer_2021, geva_transformer_2022}: We observe how the model's next token prediction is constructed throughout the layers, by de-embedding the residual stream after each attention and MLP module update (i.e., mapping it to a word), which reveals the contributions of the individual modules to the result generation process.

\begin{figure}[t]
    \includegraphics[width=0.475\textwidth]{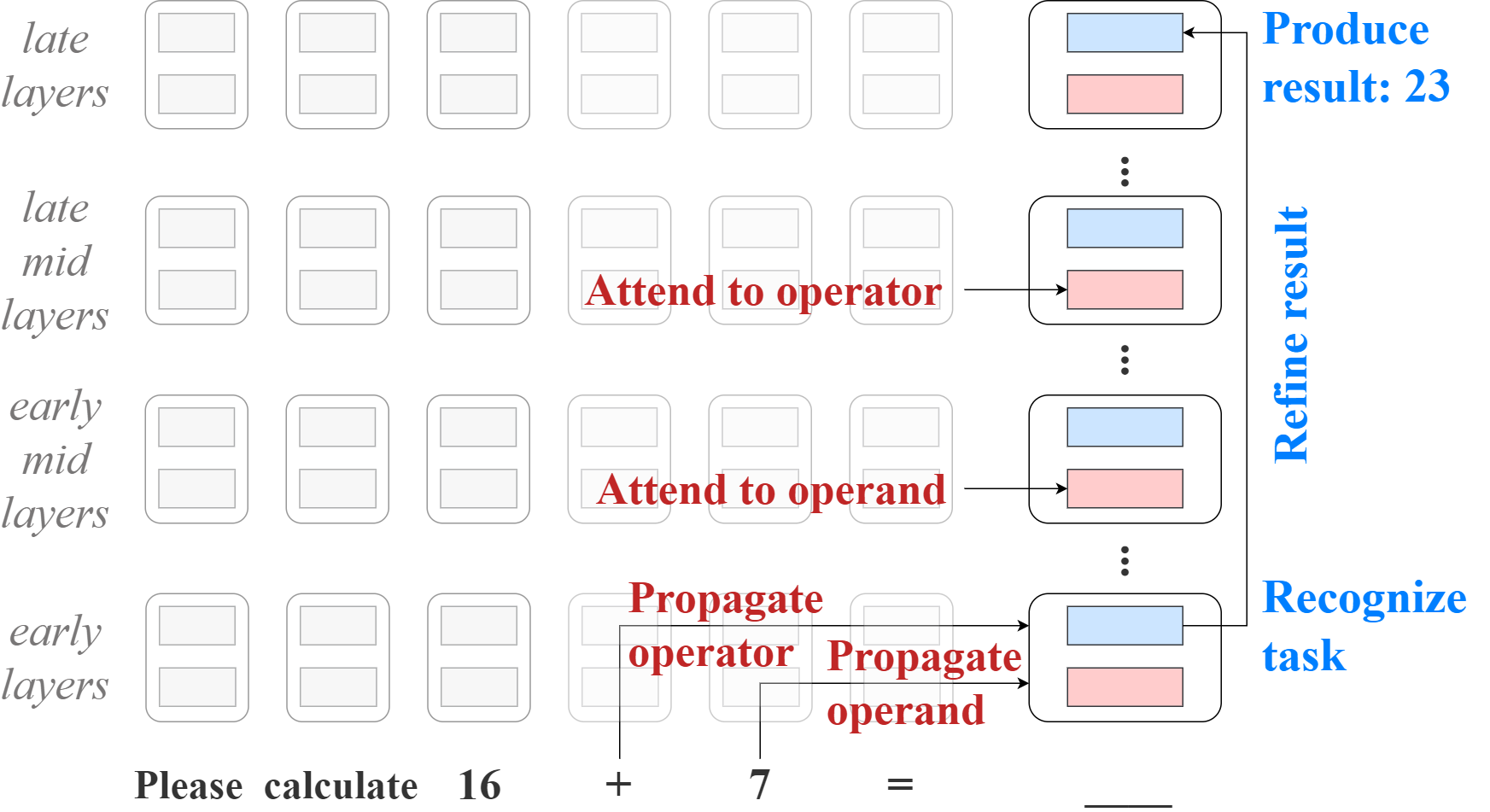}
    \caption[Visualization of the model-internal mechanisms during mathematical reasoning.]
    {Visualization of the model-internal mechanisms during mathematical reasoning. Attentions blocks are depicted in red, MLP blocks in blue.} 
    
    \label{fig:result}
\end{figure}

We present three sets of experiments to understand (1) how and when the task is recognized and the result generated, (2) how and when information from the input is propagated through the layers of the network and (3) how operands are combined, by looking at settings with 2 and 3 operands. Our main findings are, as depicted in Figure \ref{fig:result}:
\begin{itemize}
    
    \item We show that LLMs recognize the task at hand in early layers, but the generation of the correct output happens late. 
    \item We find strong indications that LLMs solve arithmetic tasks in a function-like manner, where one operand serves as an argument and the other operand is altered based on the operator and the `argument-operand', indicating processing capabilities beyond factual recall.

    \item We show that models with good arithmetic capabilities show a clear division of tasks between attention and MLP modules: Similar to previous work \cite{geva-etal-2023-dissecting}, we find that the MLP modules \textbf{aggregate} information, while the attention modules \textbf{propagate} information. This division of tasks is absent in models that struggle with arithmetic tasks.

\end{itemize}

\section{Methodology}

To investigate the mechanisms that decoder-only transformer language models use to approach and solve arithmetic tasks, we observe how the model's next token prediction is constructed throughout the layers, by projecting the residual stream after each attention and MLP module into the vocabulary space through early decoding resulting in intermediate predictions \cite{nostalgebraist, geva_transformer_2021, geva_transformer_2022}. 

This enables us to understand which token the module is currently focused on and thus
reveals the contributions of individual modules to solving the task. We analyze intermediate predictions on simple arithmetic tasks for two decoder-only language models with drastically different arithmetic capabilities, to understand what mechanisms enable models to generate correct results for arithmetic tasks.

\begin{figure}[h]

    \includegraphics[trim={0 25 0 10},width=0.475\textwidth]{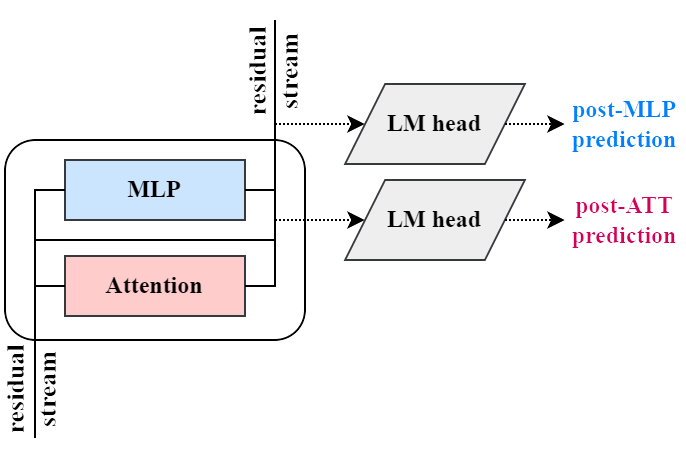}
    \caption[Visualization of early decoding. The residual stream is decoded by the LM head, once after the attention output is added to the residual stream and once after the MLP output is added.]
    {Visualization of early decoding. The residual stream is decoded by the LM head, once after the attention output is added to the residual stream (post-ATT prediction) and once after the MLP output is added (post-MLP prediction).} 
    \label{fig:method}
\end{figure}

\subsection{Early Decoding}
The method we employ for investigating the mechanisms that LLMs use to solve arithmetic tasks, was initially introduced as `logit lens' \cite{nostalgebraist}. It allows insights into how transformer-based models update the next token prediction throughout the generation process, by projecting the residual stream of the predicted token into the vocabulary space at intermediate layers.

Each attention and MLP component in a transformer-based LLM takes as input the current final token representation, i.e., the residual stream. 
in the stream, the output of each MLP and attention component is updated by adding the module's output to the input representation. 
The representation of the next token prediction at layer $k$ thus includes all the preceding additive updates that have been made to the predicted token representation by the modules in previous layers $1$ to $k$.

A language modeling head (LM head), i.e., a linear prediction layer, is used in LLMs to produce probability distributions for the next token based on the last token representation. 
This LM head can also be applied to intermediate representations from middle layers. This effectively implements a \textit{de-embedding} mechanism that allows to investigate 
the most likely intermediate prediction at each step of the generation process. Such \textit{Intermediate predictions} allow, as a consequence, to retrace the changes made to the representation by each module, and to determine the contributions of individual modules to the result generation process. The method is visualized in Figure \ref{fig:method}.

\subsection{Task and Data}
To create a controlled environment for observing LLM behavior on arithmetic tasks, we generate an artificial dataset with arithmetic tasks, which we control w.r.t.~operators (summation, denoted as \textit{add} henceforth, and subtraction, denoted as \textit{sub}), operand size, number of operands, and result size.

We prompt the models with queries of the type `Please calculate operand 1 $\circ$ operand 2 =', where operand1, operand2 $\in \mathbb{N}$ and $\circ \in \{+, -\}$, for example `Please calculate 306 + 136 =', with the correct response being `442'.

For each operator $\circ \in \{+, -\}$, we create one dataset with smaller numbers, i.e., operands and results, $add_{small}$ and $sub_{small}$, and one with larger numbers $add_{large}$ and $sub_{large}$. For the $small$ datasets, both operands and the result are $\leq 99$. In the dataset creation we ensure that all operands and results are integers $\leq 520$\footnote{We choose this upper bound because this ensures that single-token number representations are used: The vocabularies of both GPT-2 XL and GPT-NeoX-20B tokenizers contain individual tokens for all integers between 0 and 520. Higher numbers may be encoded as multiple tokens.}. Each dataset contains 500 unique queries. 
All experiments are done in a zero-shot fashion, without training or adaptation of models. 

In the remainder of the paper, we focus on an evaluation on the $add$ datasets, as we find structurally similar results for the datasets of both operands. We report differences between addition and subtraction where necessary. A full evaluation of the $sub$ datasets is provided in the Appendix.

\subsection{Models}

We experiment with two decoder-only transformer language models: GPT-NeoX-20B \cite{GPTneox2022} and GPT-2 XL \cite{radford2019languageGPT2xl}\footnote{We use the freely available \textit{EleutherAI/gpt-neox-20b} (\url{https://huggingface.co/EleutherAI/gpt-neox-20b/tree/main}) and \textit{openai-community/gpt2-xl} (\url{https://huggingface.co/openai-community/gpt2-xl}) variants on \textit{Huggingface}.}.
We confirm previous findings on GPT-2 XL's inability to solve simple arithmetic tasks, while GPT-Neox-20b performs well (Table \ref{tab:accuracy} in the Appendix). Thus, in the remainder of the paper, we focus on GPT-Neox-20b. Nevertheless, we present results on GPT-2 XL in the Appendix, as the differences in the internal mechanisms compared to GPT-Neox-20b are of interest. 

\section{Experiment Set 1: Task Recognition and Result Generation}
By investigating intermediate predictions (IP's) after MLP and ATTN modules, we first examine two fundamental questions: \textit{When does the model recognize that it needs to perform an arithmetic task?} and \textit{When does it start to generate the result?} We find answers to these questions primarily in the post-MLP predictions:
\begin{itemize}
    \item The model recognizes that it has to solve a numerical task early, and considers unspecific numerical tokens until the mid layers, where the operands are loaded (Section \ref{subsubsec:taskrecognition})
    
    \item The correct result is only predicted in the last layer (Section \ref{subsection:relativeError}).
\end{itemize}

\subsection{When is the Task Recognized?}
\label{subsubsec:taskrecognition}
\begin{figure*}[t]
    \centering
    \begin{subfigure}[b]{0.475\textwidth}
        \centering
        \includegraphics[width=\textwidth]{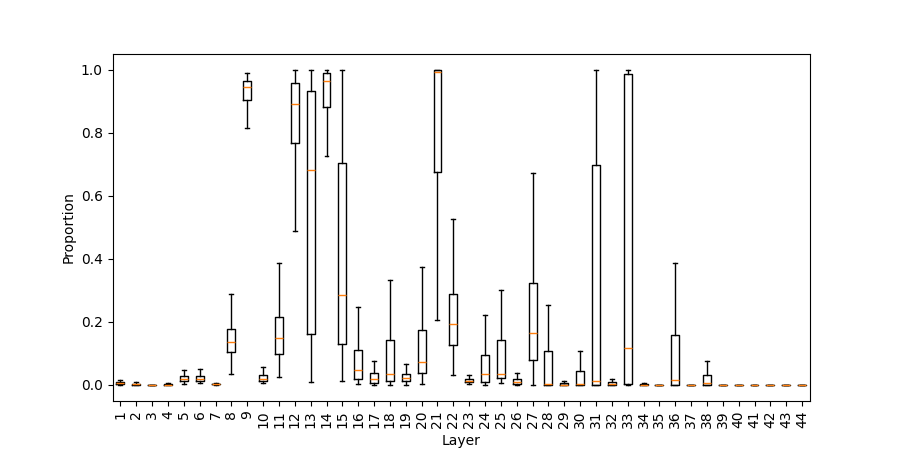}
        \caption[(a)]%
        {}    
    \end{subfigure}
    \hfill
    \begin{subfigure}[b]{0.475\textwidth}  
        \centering 
        \includegraphics[width=\textwidth]{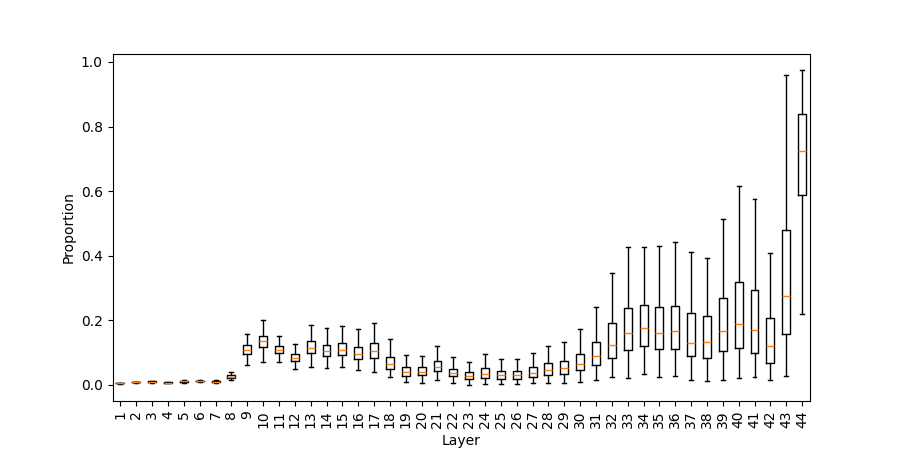}
        \caption[(b)]%
        {}   
    \end{subfigure}
    \caption[Proportion numerical addition]
    {Combined probability mass of numerical tokens in the (a) post-ATT and (b) post-MLP intermediate predictions, averaged on the $add_{large}$ dataset.} 
    \label{fig:numericalProbmass}
\end{figure*}

\paragraph{The model predicts numerical tokens early.}
We observe the probability mass of numerical tokens in the post-MLP and post-ATT IPs across different layers. 
Figure \ref{fig:numericalProbmass} shows the averaged probability mass that is assigned to numerical tokens in the $add_{large}$ dataset. There is a sharp increase in the proportion of numerical predictions around layer 9 in the post-MLP IPs, which could indicate that the model begins to recognize the numerical nature of the task. We find similar general trends in the $add_{small}$ dataset.

\begin{figure*}[t]
    \centering
    \begin{subfigure}[b]{0.475\textwidth}
        \centering
        \includegraphics[width=\textwidth]{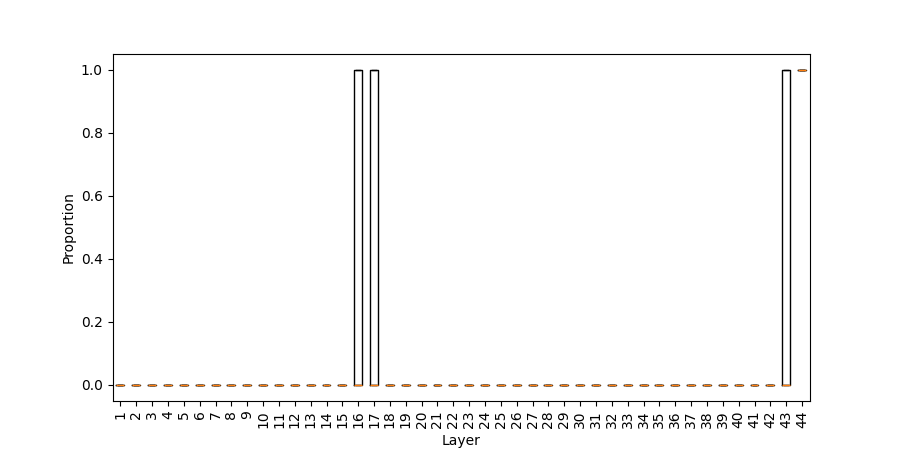}
        \caption[(a)]%
        {}
    \end{subfigure}
    \hfill
    \begin{subfigure}[b]{0.475\textwidth}  
        \centering 
        \includegraphics[width=\textwidth]{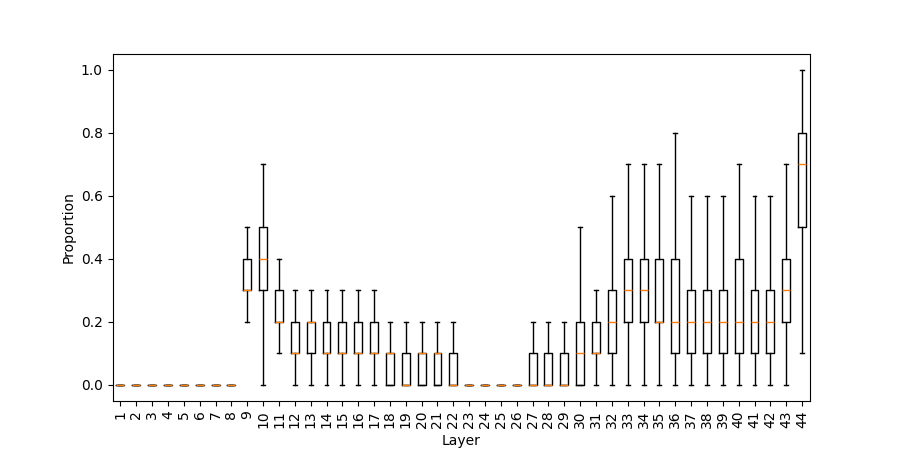}
        \caption[(b)]%
        {}   
    \end{subfigure}
    \caption[ Proportion numerical addition]
    {Proportion of numerical tokens in the (a) top 1 and (b) top 10 post-MLP intermediate predictions, averaged on the $add_{large}$ dataset.} 
    \label{fig:numericalproportion}
\end{figure*}

\paragraph{Numerical tokens are only predicted with high confidence in the last layer.}
Figure \ref{fig:numericalproportion} shows the average proportion of numerical tokens within the top 1 and top 10 IPs, on the $add_{large}$ dataset. The number of numerical predictions in the top 10 predicted tokens post-MLP is between 10 and 40\% for early-mid layers (layers 9 to 17) and for mid-late layers (layers 30 to 43). However, the model only predicts a numerical token as the top prediction in the very last layer, i.e., layer 44. 
\paragraph{The behavior of attention layers differs significantly.}
The patterns observed for the numerical predictions after the attention modules are very different. We observe significant spikes in numerical predictions at specific layers, particularly at layers 9, 12, 14, and 21, where 87\% to 99\% of the probability mass is on numerical tokens on average. Previous work has shown that attention modules are responsible for propagating information between positions \cite{geva-etal-2023-dissecting}, thus we conjecture that these spikes indicate that important numerical information may be attended to in the final token position or propagated to the final token position via the prominent attention modules. We look into these findings in more detail in Section \ref{subsubsec:operands}.


\subsection{When is the Correct Result Generated?}
\label{subsection:relativeError}
\paragraph{Numerical predictions are unrelated to the correct result until late layers.}

\begin{figure*}[t]
    \centering
    \begin{subfigure}[b]{0.475\textwidth}
        \centering
        \includegraphics[width=\textwidth]{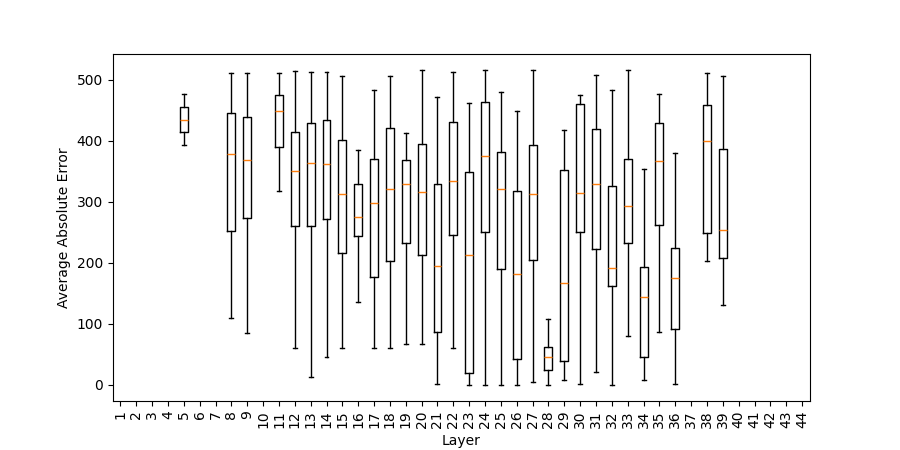}
        \caption[(a)]%
        {}   
    \end{subfigure}
    \hfill
    \begin{subfigure}[b]{0.475\textwidth}  
        \centering 
        \includegraphics[width=\textwidth]{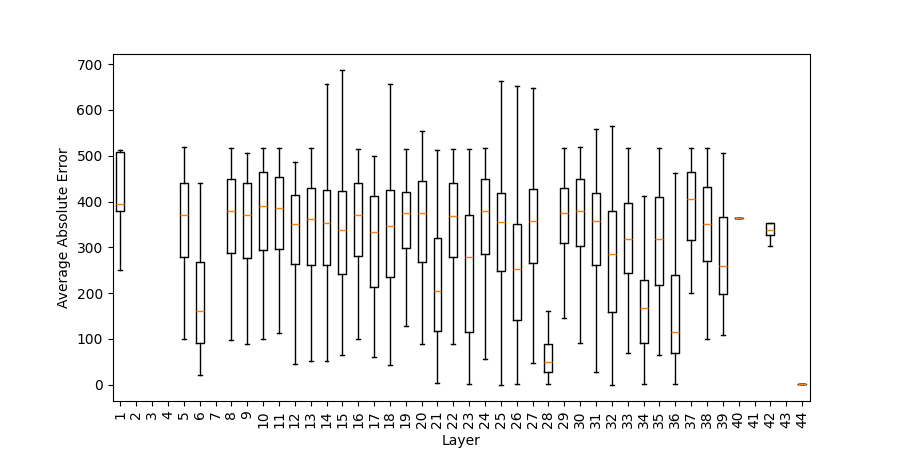}
        \caption[(b)]%
        {}  
    \end{subfigure}
    \vskip 0.5\baselineskip
    \begin{subfigure}[b]{0.475\textwidth}   
        \centering 
        \includegraphics[width=\textwidth]{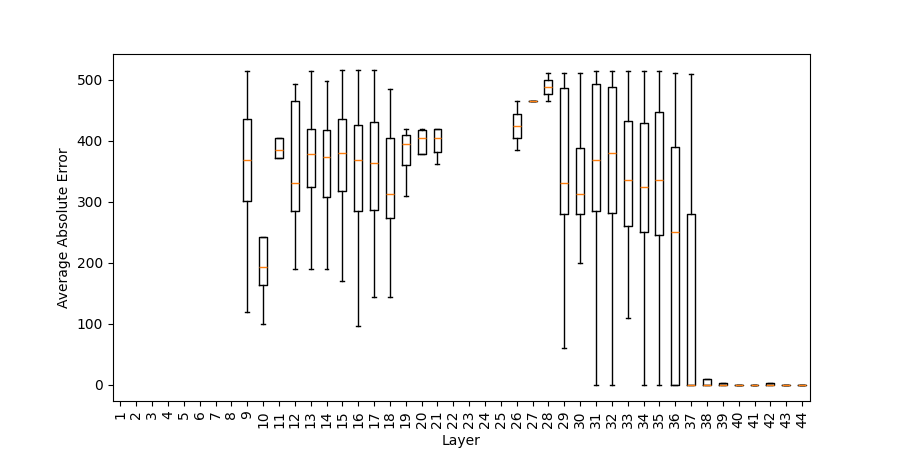}
        \caption[(c)]%
        {}   
    \end{subfigure}
    \hfill
    \begin{subfigure}[b]{0.475\textwidth}   
        \centering 
        \includegraphics[width=\textwidth]{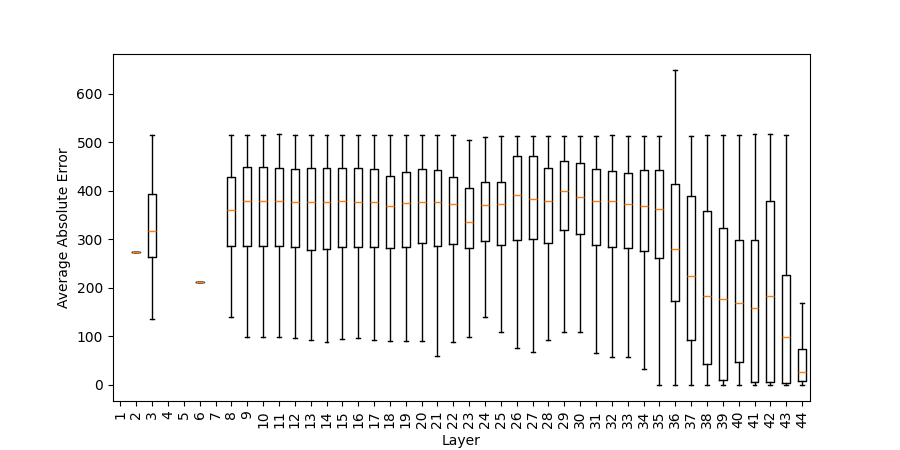}
        \caption[(d)]%
        {}  
    \end{subfigure}
    \caption[ absolute error numerical addition]
    {Absolute error, i.e., difference to correct result, of numerical tokens in the (a) top 1 and (b) top 10 post-ATT intermediate predictions. And absolute error of numerical tokens in the (c) top 1 and (d) top 10 post-MLP intermediate predictions. All averaged on the $add_{large}$ dataset.} 
    \label{fig:absoluteErrorPredictions}
\end{figure*}

\begin{figure*}[t]
    \centering
    \begin{subfigure}[b]{0.475\textwidth}   
        \centering 
        \includegraphics[width=\textwidth]{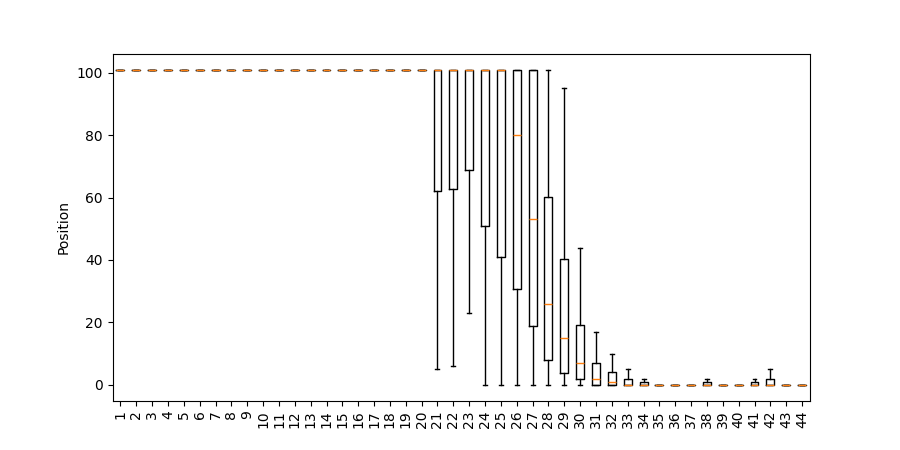}
        \caption[(a)]%
        {}    
    \end{subfigure}
    \hfill
    \begin{subfigure}[b]{0.475\textwidth}   
        \centering 
        \includegraphics[width=\textwidth]{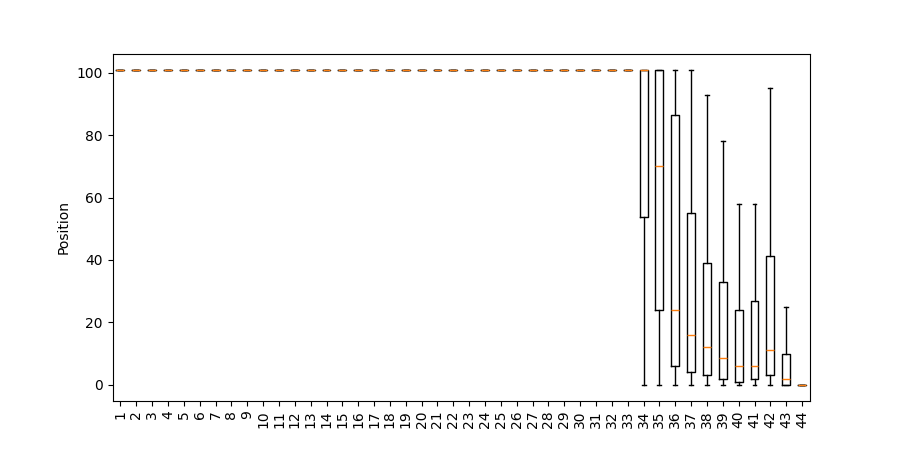}
        \caption[(b)]%
        {}    
    \end{subfigure}
    \caption[ Correct result position ]
    {Position of correct result in the post-MLP prediction of intermediate layers, averaged over all data points in (a) $add_{small}$ and (b) $add_{large}$.} 
    \label{fig:correct_result}
\end{figure*}

To understand when the result is generated, we examine the similarity of numerical IPs to the correct result throughout the layers. 
We analyze the absolute error, defined as the difference between the predicted number and the correct number, within the top 10 and top 1 predicted tokens (Figure \ref{fig:absoluteErrorPredictions}). 
Our analysis reveals several key findings: Firstly, in the post-MLP IPs, the similarity between the predicted numerical tokens and the correct result is generally low in early and mid layers, but begins to increase incrementally after layer 35, corresponding to the layer where the correct result is assigned a higher probability (Figure \ref{fig:correct_result}). There is a notable decrease in absolute error for the top 1 and top 10 post-ATT predictions in layer 28, indicating that the model has a reasonable approximation of the magnitude of the correct result at that layer.

\paragraph{The correct result is produced only in very late layers.} We also analyze at which layer the correct result is produced and evaluate both the probability of the result token and its position among the most probable tokens at each layer.

The correct result typically emerges in the later layers of the model (Figure \ref{fig:correct_result}). This is demonstrated by a significant increase in the correct token's probability and a corresponding decrease in its position, with the correct result appearing around layers 35-40 for large addition tasks, with a final sharp increase of probability and decrease of rank in the final layer 44. For smaller addition tasks the correct result token emerges as early as layer 26 and is anchored as the highest predicted token around layer 32-34. The correct result thus appears earlier in the generation process of easier tasks compared to more complex ones. This could indicate different internal mechanisms for solving easier compared to more challenging mathematical reasoning tasks.

\section{Experiment Set 2: Input Propagation}
Our findings in the previous section provide insights into where in the model the task is recognized and the result emerges. However, the specific mechanisms underlying the result generation remain unclear. To address this gap, we examine how and when critical information from the input - necessary for solving the arithmetic task - is propagated through the model.

Specifically, we answer two fundamental questions: \textit{How and when is operand information propagated?} and \textit{How and when is operator information propagated?}. We find that:
\begin{itemize}
    \item Mid-layer attention modules propagate one or both operands into the residual stream 
    (Section \ref{subsubsec:operands}).
    \item Mid- to late-layer attention modules 
    propagate the operator into the residual stream 
    (Section \ref{subsec:taskmarkers}).

\end{itemize}

\subsection{How and when is operand information propagated?}
\label{subsubsec:operands}

\paragraph{Specific attention modules propagate numerical information.} Figure \ref{fig:numericalProbmass} shows that the post-ATT predictions display clear spikes in numerical predictions at specific layers. The most prominent spikes occur at layers 9, 12, 14, and 21, where around 87\% to 99\% of the probability mass is distributed on numerical tokens on average. We conjecture that these spikes could be a trace of important numerical information that is propagated to the final token position via these attention modules. We hypothesize that this information may include the operands required for solving the arithmetic task. To explore this further, we examine the numerical tokens predicted in the post-ATT IPs.

\paragraph{Operands are explicitly propagated to the final token position.} Given that attention modules have been shown to propagate relevant input information to the final prediction token \cite{geva-etal-2023-dissecting}, we investigate whether attention modules propagate operand information when solving arithmetic tasks.

We thus examine whether operands are predicted with high probability in the post-ATT IPs across layers. We find that both operands are predicted as the highest probability (rank 1) token, in 45–49\% of queries. Specifically, operand 1 is the highest predicted token at the post-ATT representation of \textit{some} layer (mean layer 21.4) in 49\% of cases, and operand 2 (mean layer 21.0) in 45\%. This indicates that operands are in fact propagated to the final token position.

Neither operand appears in the top predictions after any MLP module 
, further supporting the evidence that the model only propagates the operands to the final token via the attention modules. 

We find similar trends of operand propagation for negation, see Appendix \ref{sec:sub_appendix}.

\begin{figure*}[t]
    \centering
    \begin{subfigure}[b]{0.475\textwidth}
        \centering
        \includegraphics[width=\textwidth]{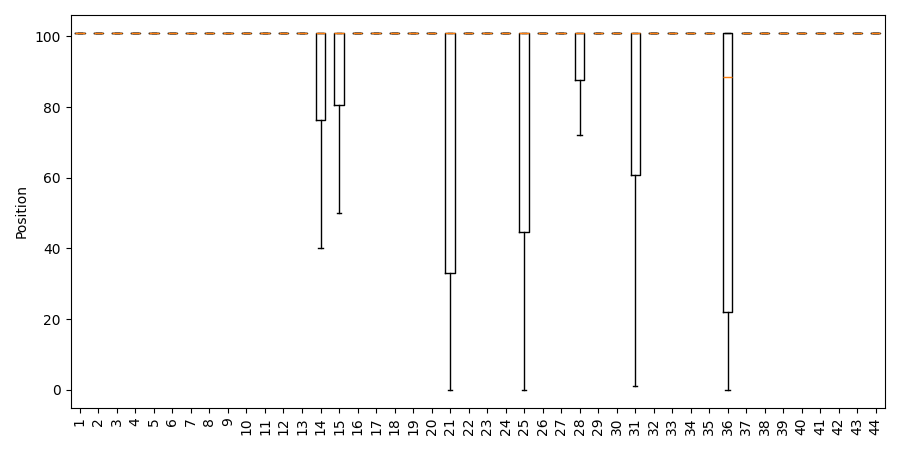}
        \caption[(a)]%
        {}    
    \end{subfigure}
    \hfill
    \begin{subfigure}[b]{0.475\textwidth}  
        \centering 
        \includegraphics[width=\textwidth]{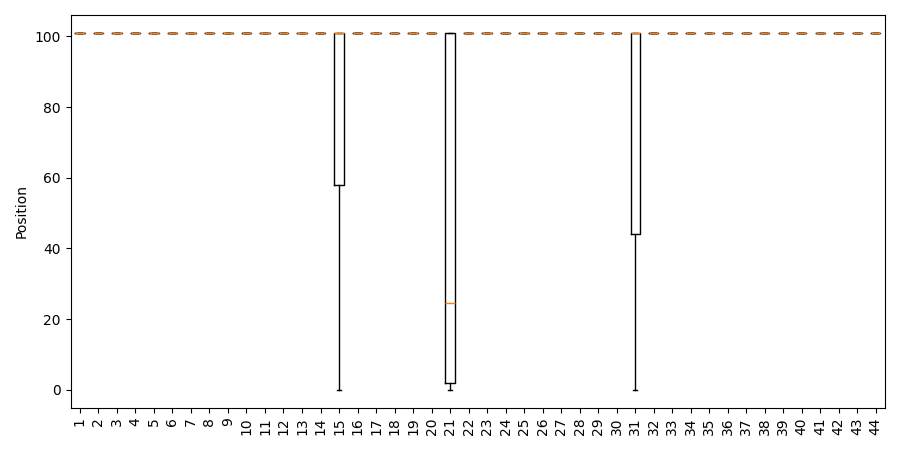}
        \caption[(b)]%
        {}    
    \end{subfigure}
    \caption[ Operands 'loaded' by the attention layer on the final token ]
    {Position of (a) operand 1 and (b) operand 2 in the post-ATT prediction of intermediate layers, averaged over all data points in $add_{large}$.} 
    \label{fig:operands}
\end{figure*}

\paragraph{The model mostly only propagates one of the operands.} We find that in 64\% of cases, the propagation of operands is \textit{mutually exclusive}, such that when operand 1 occurs as the top prediction in a post-ATT IP, operand 2 does not, and vice versa. This suggests that the model may only need to explicitly attend to one of the operands of the arithmetic task, to accurately solve it.
Since the information from both operands is required to predict an accurate result, we would have expected that both operands would be attended to during the next token prediction. We thus conjecture that one of the operands must be represented in the residual stream of the next token prediction in a more implicit way.
This finding suggests that the model may be operating in a function-like manner, where one operand is explicitly propagated as the argument, while the other operand is implicitly represented and iteratively transformed into the result.

\subsection{How and when is operator information propagated?}
\label{subsec:taskmarkers}

To investigate when and where operators are propagated, we look through all tokens that are consistently predicted with a high probability, i.e., an average probability of >0.5 in >80\% of tokens.

\begin{table}[htbp]
\centering
\begin{tabular}{|c|c|c|c|}
\hline
Layer & Token & Mean Prob. & Frequency \\
\hline
10  & ` \textbackslash n'  & 0.62 & 90.60\% \\
35 & ` \textbackslash n'  & 0.94 & 81.20\% \\
37 & ` +'  & 0.99 & 97.60\% \\
43 & ` answ'  & 0.78 & 88.60\% \\
\hline
\end{tabular}
\caption{Frequent, high probability tokens predicted post-ATT, in the $add_{large}$ dataset.}
\label{tab:otherAddLarge}
\end{table}

We only find such tokens in the post-ATT, but not the post-MLP representation. 
We first find that the operator is reliably propagated to the final token position  at layer 37 with very high probability (> 0.98) (Table \ref{tab:otherAddLarge}). Results for the other data sets can be found in the Appendix. 
Around the same layer is where the token corresponding to the correct result starts to get considered as high probability prediction (Figure \ref{fig:correct_result}). 
Another interesting observation is the consistent generation of the token `answ' in the post-ATT IP of the second to last layer. Since the correct result is only predicted with high probability in the last layer (layer 44) (Figure \ref{fig:numericalproportion}), it is conceivable that the previous layer's attention module is responsible for prompting the final MLP module to bump up the probability of the iteratively generated result. The reliably generated token in the post-ATT IP of layer 43 might be an indication for this.

\section{Experiment Set 3: More Operands}
The model seems to be able to correctly solve an arithmetic task with two operands and one operator, without explicitly attending to all three pieces of information in the residual stream of the final token (Section \ref{subsubsec:operands}).
Given that it should not be possible to construct the correct result for a task without loading both operands and the operator, the `missing operand' has to be loaded into the representation of the final token in a more implicit way, which we cannot observe with our method.

To test this intuition, we perform an ablation study with 3 operands and 2 operators. If our intuition is correct, we should find that the model is able to successfully solve the task if at least 2 of the operands and both operators are explicitly propagated to the final token representation. It should fail if only one of the operands, or only one of the operators is explicitly attended to.

We generate 4 datasets of 100 samples of the form `Please calculate $op 1 \circ op 2$ {\scriptsize $\square$} $op 3$ =', where $op 1, op 2, op 3 \in \mathbb{N}$, $result < 520$ and $\circ,$ {\scriptsize $\square$} $\in \{+, -\}$, for example `Please calculate 75 + 16 - 48 ='. Each dataset contains one combination of $\circ$ and {\scriptsize $\square$}, e.g., (+, -).

The performance of GPT-Neox-20b is very poor on these datasets (accuracy $\leq$ 0.12 on all datasets). 
We investigate whether this may be due to insufficient operand information being propagated to the final token position, by observing how many operands are generated as the top prediction in the post-ATT prediction of any layer. And in fact, we find that in 62\% of samples across datasets the propagated operand information is not sufficient to generate the correct result, i.e., only one or none of the operands are propagated. In addition, neither the (+, -) nor the (-, +) dataset generates both operators as post-ATT predictions, further indicating that the model's failure on 3-operand tasks may be due to insufficient input information being incorporated into the result generation process.
\section{Interventions on Operands}
To verify that the operands are in fact propagated to the final token via specific attention modules, we perform interchange interventions on the attention outputs of the final token position.

Using the pyvene library \cite{wu_pyvene_2024}, we systematically intervene on attention modules of the last input token by interchanging the attention outputs of individual layers between a base and a source query. Specifically, during the forward pass on the base query, the output of an attention module is replaced with the output of that attention module on the source query.
We use source queries that differ from the base query in exactly one operand or operator, e.g., $base$ := `Please calculate 78 + 62 =' and $source_{operand2-intervention}$ := `Please calculate 78 + 93 =', and $source_{operator-intervention}$ := `Please calculate 78 - 62 ='. 

\begin{figure*}[t]
    \centering
    \includegraphics[width=.9\textwidth]{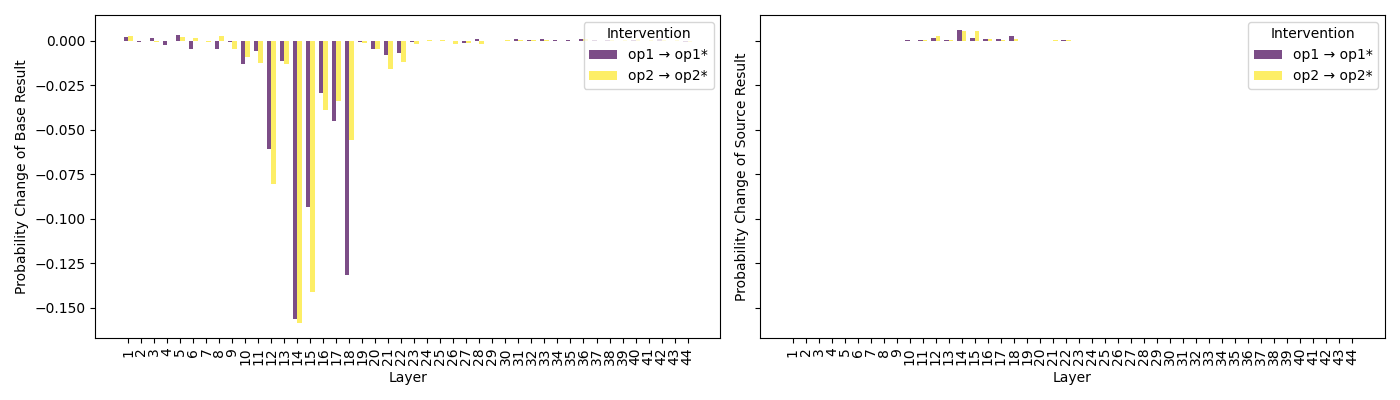}
    \caption[Intervention operands]
    {Effect of interchange intervention, i.e., source is used to intervene on base, on one of the operands in the $add_{large}$ dataset. 
    Left: Decrease in probability of base result in early-mid layers. Right: Change in probability of source result in early-mid layers.} 
    \label{fig:interventionOperands}
\end{figure*}

\begin{figure*}[t]
    \centering
    \includegraphics[width=.9\textwidth]{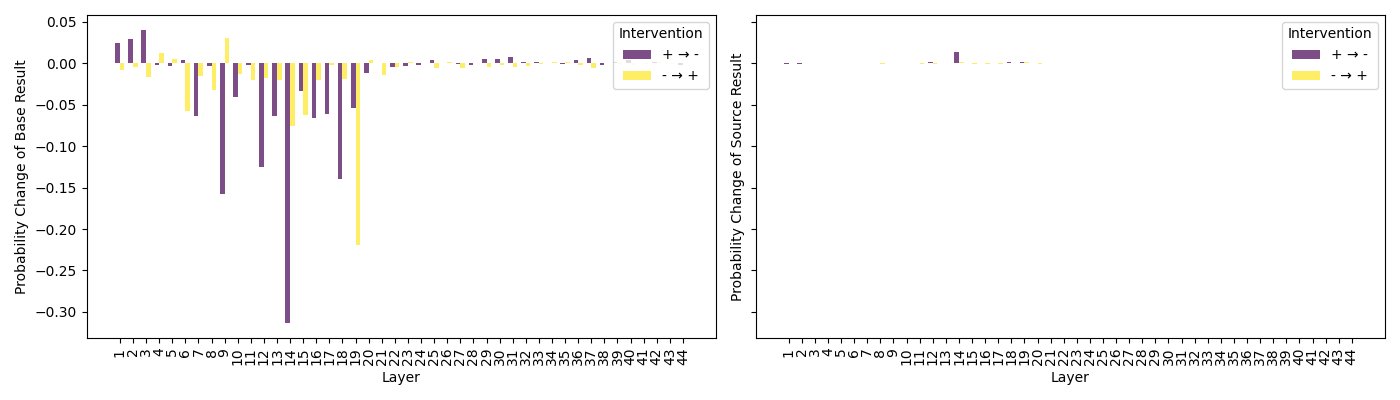}
    \caption[Intervention operators]
    {Effect of interchange intervention, i.e., source is used to intervene on base, on the operator in the $add_{large}$ dataset (purple bars) and the $sub_{large}$ dataset (yellow bars). 
    Left: Decrease in probability of base result in early-mid layers. Right: Change in probability of source result in early-mid layers.} 
    \label{fig:interventionOperator}
\end{figure*}

By observing the effect of the intervention on the predicted token we can quantify the importance of individual attention modules for propagating the manipulated part of the input. Specifically, we observe the change in prediction probability of the base-result and the source-result.
Figure \ref{fig:interventionOperands} and figure \ref{fig:interventionOperator} show that the model is very sensitive to the intervention on specific parts of the input, e.g., removing operator (`+') information on the final input position in layer 14 drastically decreases the probability of the correct (`base') result by more than 30\%.  
In accordance with the insights from the IP investigations, we conjecture that early- to mid-layer attention modules (layer 12 to 19) are responsible for propagating input information into the residual stream of the final token, while subsequent attention modules (starting around layer 20) attend to this information (Section \ref{subsubsec:operands} so that it can be used in the result generation process by the MLP modules.

\section{Related Work}
\label{sec:related-work}

Arithmetic tasks have become a key benchmark for assessing the mathematical reasoning abilities of large language models (LLMs). Despite scaling laws suggesting that LLMs performance increases with model size and training data \cite{kaplan2020scaling,hoffmann2022training}, LLMs struggle to reliably generate accurate arithmetic results. A growing body of research evaluates the accuracy of LLMs on tasks involving arithmetic reasoning \cite{shakarian2023independent,fu2023chain,hong2024stuck,frieder2024mathematical} as well as their robustness \cite{anantheswaran2024investigatingrobustnessllmsmath,stolfo_causal_2023,wei2022chain,mishra_numglue_2022,cobbe2021training}. Besides the evaluation on simple arithmetic tasks, Math Word Problem (MWP) datasets \cite{koncel2016mawps,patel2021nlp,miao2020diverse,hosseini-etal-2014-learning,hendrycks2021measuringA,frieder2024mathematical} provide benchmarks of varying difficulty for models' ability to recognize and solve mathematical reasoning tasks.

However there is limited work that investigates \textit{how} LLMs generate results of arithmetic tasks. In this paper we turn to techniques from mechanistic interpretability to investigate the arithmetic processing in LLMs. Mechanistic interpretability aims at reverse engineering the algorithms that are hypothesized to be encoded in the model parameters (\citealt{olah2020zoom}). Methods from mechanistic interpretability have since led to detailed insights on how information is processed by state-of-the-art language models \cite{geva_transformer_2021,geva-etal-2023-dissecting,meng_locating_2023,nanda_emergent_2023,olsson_-context_2022}. 

Notably, causal mediation analysis \cite{vig_investigating_2020,meng_locating_2023} has been applied to investigate arithmetic reasoning in LLMs \cite{stolfo_mechanistic_2023}. This work reveals that attention mechanisms within LLMs transmit relevant information in the input to the final token, where MLP modules are responsible for generating the correct result. 

However, causal mediation analysis cannot provide a detailed understanding of how the final token is computed. It only identifies modules capable of producing the correct output and is not suited for investigating possible intermediate steps. 

Recent work provides complementary perspectives on arithmetic processing in LLMs.
While some studies suggest that arithmetic is solved through heuristics or pattern recognition rather than coherent algorithms \cite{nikankin2024arithmetic, deng2024language}, others highlight more structured, algorithmic mechanisms, such as digit-wise or position-specific pathways \citep{lindsey2025biology, baeumel2025modulararithmeticlanguagemodels, baeumel2025lookaheadlimitationmultioperandaddition}. Despite these advances, a unified understanding of how LLMs execute arithmetic remains open.

In this paper, we take a closer look at how LLMs generate results by inspecting intermediate model predictions on the final token. 
Prior work shows that MLP layers in transformers incrementally build predictions by promoting concepts that are interpretable in the vocabulary space \citep{geva_transformer_2021, geva_transformer_2022, geva2022lmdebuggerinteractivetoolinspection}. This method has been applied to tasks such as factual recall, where LLMs solve problems via simple vector arithmetic operations within MLP modules \citep{merullo_language_2024}. Using this approach, we aim to shed light on the intermediate steps of arithmetic processing in LLMs.
\section{Discussion \& Conclusion}
In this study, we explore the internal mechanisms of GPT-NeoX-20b 
in solving arithmetic tasks, focusing on how attention and MLP modules update the next token prediction throughout the layers. 
We find that while the model recognizes the task early, it only begins refining the predicted token in later layers, after relevant input information has been propagated.
Our analyses reveal that operands and operators are explicitly propagated into the final token position by specific attention modules. The model then iteratively generates the correct result, which is only predicted with high probability in the final layer. This highlights the critical role of attention mechanisms in propagating numerical information through the network. Interestingly, only one operand is explicitly propagated, while the other is encoded in a more implicit form, suggesting that arithmetic reasoning involves systematic transformations, similar to applying a function.
We also observe a clear division of labor between attention and MLP modules in GPT-NeoX-20b, a distinction absent in GPT-2 XL, which struggles with arithmetic tasks. This division appears to be crucial for successful arithmetic reasoning in transformer models. These findings suggest that future research should further investigate this division of labor, particularly for tasks requiring more complex reasoning, such as multi-step arithmetic calculations.

\section*{Limitations}
In our evaluation, we focus on two models, GPT-NeoX-20b and GPT-2 XL, the latter of which struggles with basic arithmetic tasks. While this contrast provides insights into the requirements for arithmetic reasoning in LLMs, it should be explored whether our findings generalize to other state of the art language models.\\
Additionally, our experiments target simple arithmetic queries, which are highly controlled and allow for straightforward analysis. However, real-world arithmetic tasks, such as solving math word problems, involve more complex reasoning and natural language understanding. Future research could explore whether the mechanisms identified for simple arithmetic tasks are comparable in more natural language prompts.\\
Finally, future work should investigate whether intervening on operands and operators generated in intermediate post-ATT predictions could significantly affect the model prediction. While our analyses clearly show that attention modules attend to this input information in the output representation, further experiments are needed to confirm that subsequent MLP modules rely on this attention.

\section*{Acknowledgements}
We thank the anonymous reviewers for their helpful feedback on the paper draft. This work has been supported by the German Federal Ministry of Research, Technology and Space (BMFTR) as part of the project TRAILS (01IW24005).

\bibliography{custom1}

\appendix
\section{Arithmetic capabilities of models}
\begin{table}[h]
\small
\centering
\begin{tabular}{|l|c|c|c|c|}
\hline
Model & $add_{S}$ & $add_{L}$ & $sub_{S}$ & $sub_{L}$ \\
\hline
GPT-Neox-20b & 0.97 & 0.79 & 1.0 & 0.95 \\
\hline
GPT-2 XL & 0.08 & 0.0 & 0.06 & 0.0 \\
\hline
\end{tabular}
\caption{Task accuracy of both models on small (S) and large (L) datasets.}
\label{tab:accuracy}
\end{table}
\section{Tokens with a High average IP Probability}

\begin{table}[htbp]
\centering
\begin{tabular}{|c|c|c|c|}
\hline
Layer & Token & Mean Prob. & Frequency \\
\hline
36  & ` +'  & 0.98 & 90.40\% \\
40 & `."'  & 0.86 & 80.00\% \\
41 & ` \$'  & 0.95 & 85.40\% \\
42 & ` answ'  & 0.89 & 100\% \\
\hline
\end{tabular}
\caption{$\mathbf{add}_{small}$: Frequent, high probability tokens predicted post-ATT}
\label{tab:otherAddSmall}
\end{table}

\begin{table}[htbp]
\centering
\begin{tabular}{|c|c|c|c|}
\hline
Layer & Token & Mean Prob. & Frequency \\
\hline
30 & ` -'  & 0.99 & 86.20\% \\
39 & ` using'  & 0.99 & 97.80\% \\
42 & ` answ'  & 0.73 & 88.40\% \\
\hline
\end{tabular}
\caption{$\mathbf{sub}_{large}$: Frequent, high probability tokens predicted post-ATT}
\label{tab:otherSubLarge}
\end{table}

\begin{table}[htbp]
\centering
\begin{tabular}{|c|c|c|c|}
\hline
Layer & Token & Mean Prob. & Frequency \\
\hline
9  & ` \textbackslash n'  & 0.67 & 86.40\% \\
30 & ` -'  & 0.99 & 93.60\% \\
39 & ` using'  & 0.98 & 84.00\% \\
42 & ` answ'  & 0.90 & 100\% \\
\hline
\end{tabular}
\caption{$\mathbf{sub}_{small}$: Frequent, high probability tokens predicted post-ATT}
\label{tab:otherSubSmall}
\end{table}
\section{Intermediate predictions on $\mathbf{sub}_{large}$}
\label{sec:sub_appendix}
\begin{figure*}[t]
    \centering
    \begin{subfigure}[b]{0.475\textwidth}
        \centering
        \includegraphics[width=\textwidth]{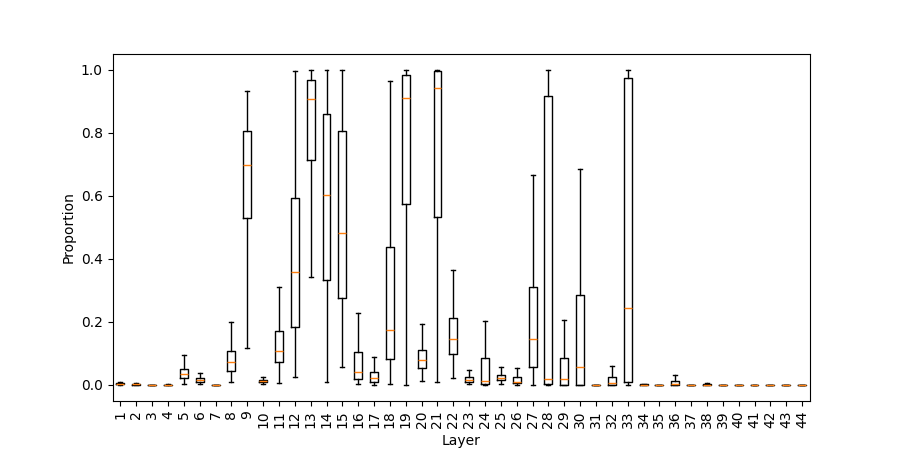}
        \caption[(a)]%
        {}    
    \end{subfigure}
    \hfill
    \begin{subfigure}[b]{0.475\textwidth}  
        \centering 
        \includegraphics[width=\textwidth]{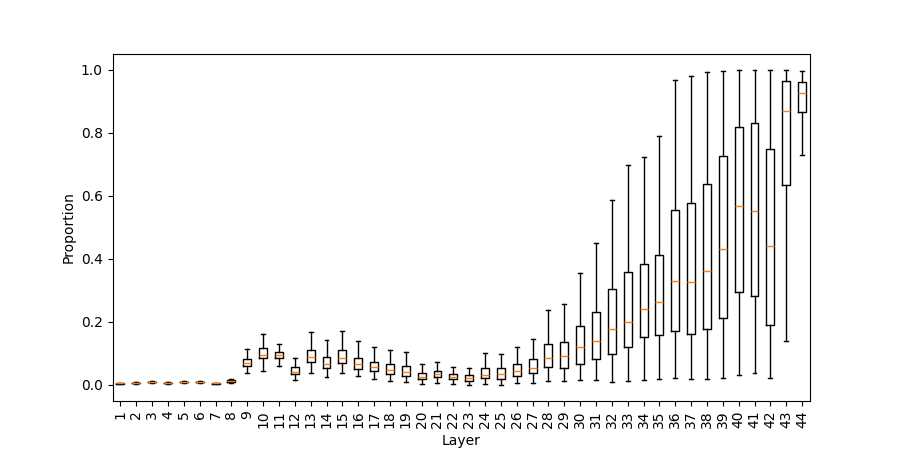}
        \caption[(b)]%
        {}   
    \end{subfigure}
    \caption[Proportion numerical subtraction]
    {GPT-Neox-20b: Combined probability mass of numerical tokens in the (a) post-ATT and (b) post-MLP intermediate predictions, averaged on the $sub_{large}$ dataset.} 
    \label{fig:numericalProbmass_sub}
\end{figure*}

\begin{figure*}[t]
    \centering
    \begin{subfigure}[b]{0.475\textwidth}
        \centering
        \includegraphics[width=\textwidth]{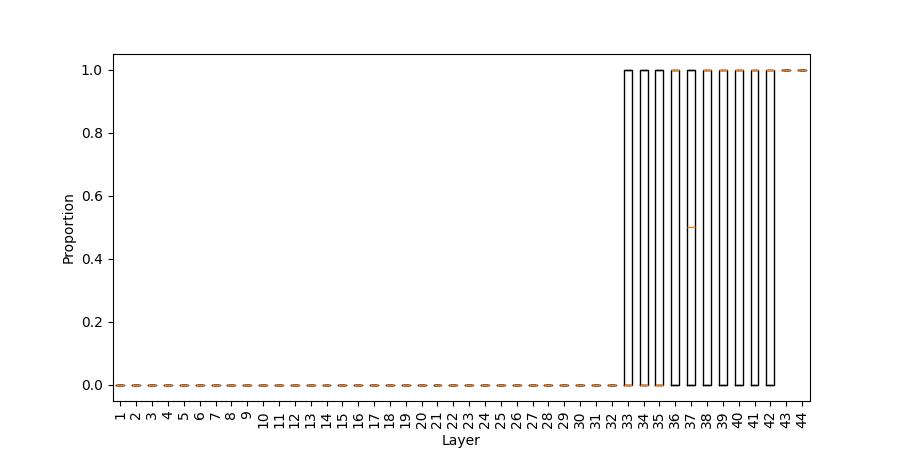}
        \caption[(a)]%
        {}
    \end{subfigure}
    \hfill
    \begin{subfigure}[b]{0.475\textwidth}  
        \centering 
        \includegraphics[width=\textwidth]{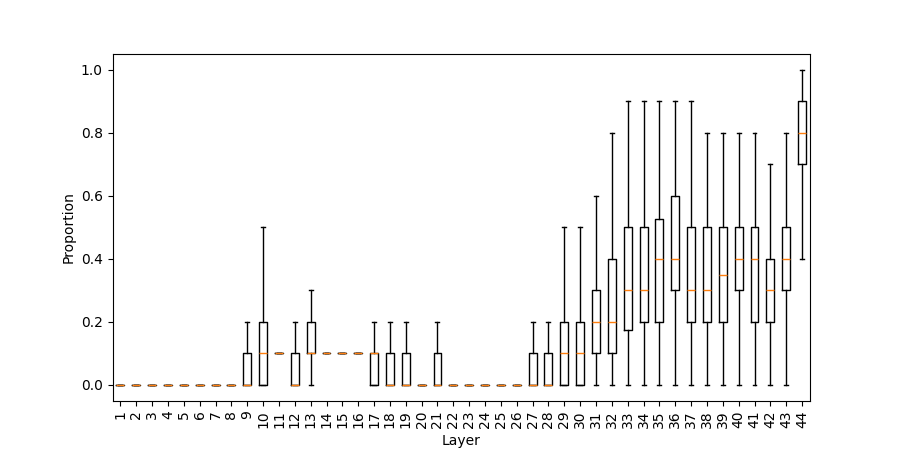}
        \caption[(b)]%
        {}   
    \end{subfigure}
    \caption[ Proportion numerical subtraction]
    {GPT-Neox-20b: Proportion of numerical tokens in the (a) top 1 and (b) top 10 post-MLP intermediate predictions, averaged on the $sub_{large}$ dataset.} 
    \label{fig:numericalproportion_sub}
\end{figure*}

\begin{figure*}[t]
    \centering
    \begin{subfigure}[b]{0.475\textwidth}
        \centering
        \includegraphics[width=\textwidth]{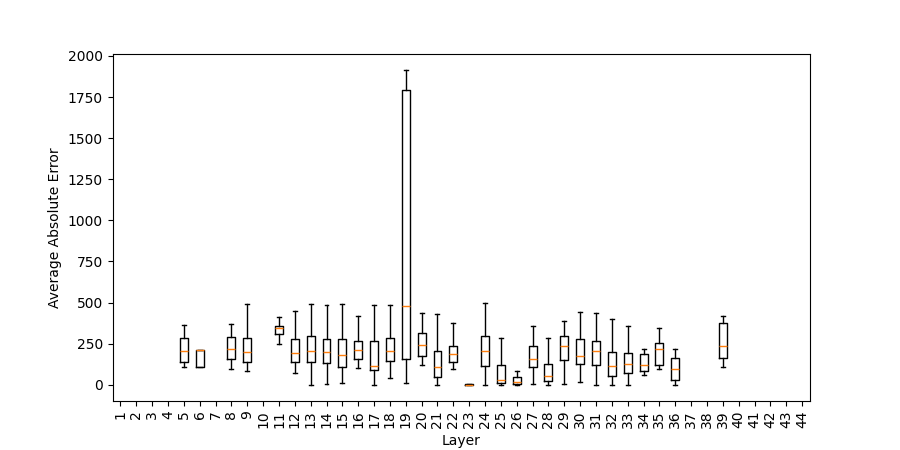}
        \caption[(a)]%
        {}   
    \end{subfigure}
    \hfill
    \begin{subfigure}[b]{0.475\textwidth}  
        \centering 
        \includegraphics[width=\textwidth]{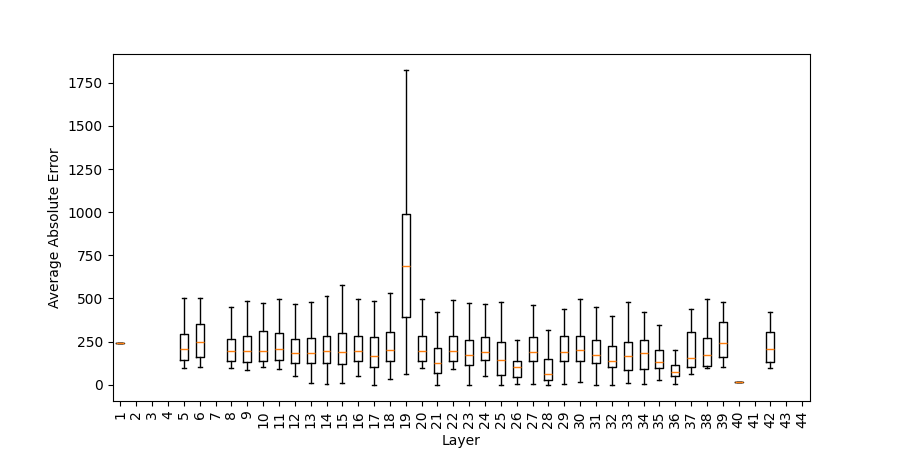}
        \caption[(b)]%
        {}  
    \end{subfigure}
    \vskip 0.5\baselineskip
    \begin{subfigure}[b]{0.475\textwidth}   
        \centering 
        \includegraphics[width=\textwidth]{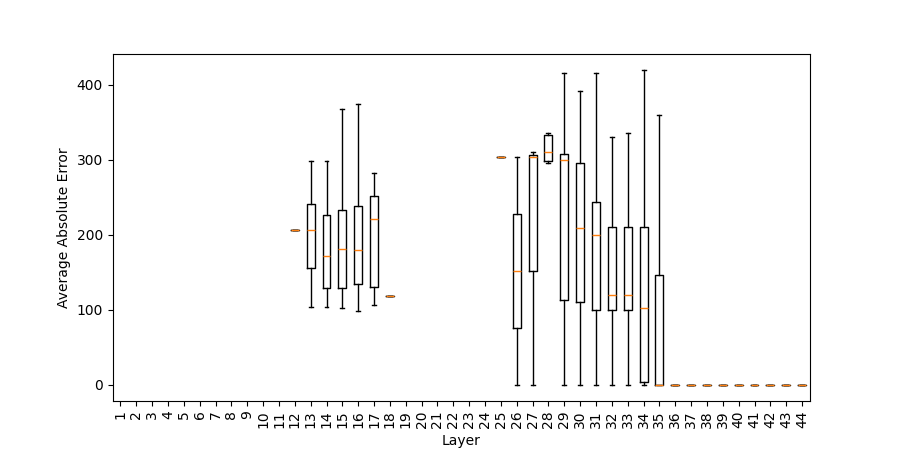}
        \caption[(c)]%
        {}   
    \end{subfigure}
    \hfill
    \begin{subfigure}[b]{0.475\textwidth}   
        \centering 
        \includegraphics[width=\textwidth]{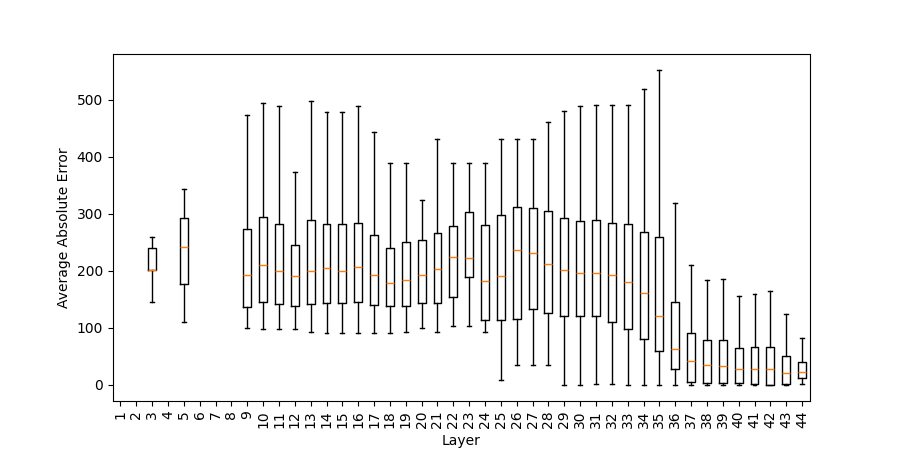}
        \caption[(d)]%
        {}  
    \end{subfigure}
    \caption[ absolute error numerical subtraction]
    {GPT-Neox-20b: Absolute error, i.e., difference to correct result, of numerical tokens in the (a) top 1 and (b) top 10 post-ATT intermediate predictions. And absolute error of numerical tokens in the (c) top 1 and (d) top 10 post-MLP intermediate predictions. All averaged on the $sub_{large}$ dataset.} 
    \label{fig:absoluteErrorPredictions_sub}
\end{figure*}

\begin{figure*}[t]
    \centering
    \begin{subfigure}[b]{0.475\textwidth}   
        \centering 
        \includegraphics[width=\textwidth]{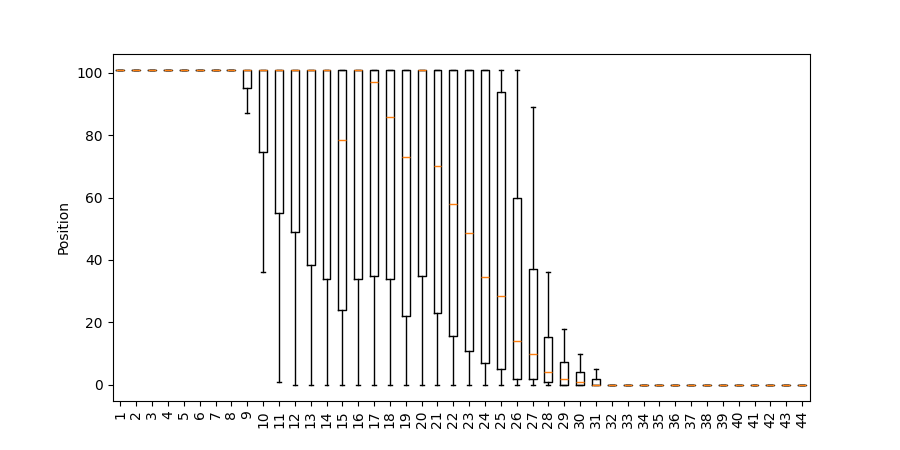}
        \caption[(a)]%
        {}    
    \end{subfigure}
    \hfill
    \begin{subfigure}[b]{0.475\textwidth}   
        \centering 
        \includegraphics[width=\textwidth]{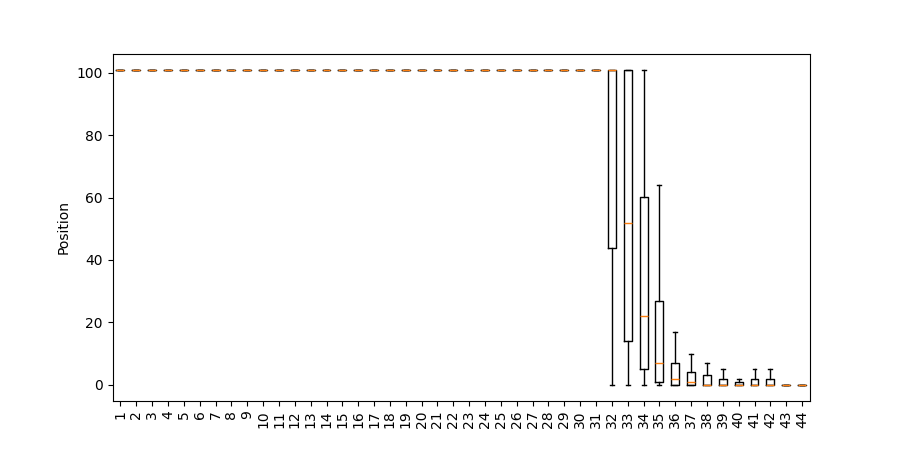}
        \caption[(b)]%
        {}    
    \end{subfigure}
    \caption[ Correct result position ]
    {GPT-Neox-20b: Position of correct result in the post-MLP prediction of intermediate layers, averaged over all data points in (a) $sub_{small}$ and (b) $sub_{large}$.} 
    \label{fig:correct_result_sub}
\end{figure*}

\paragraph{The model recognizes the task early.} 
Figure \ref{fig:numericalProbmass_sub} shows the averaged probability mass that is assigned to numerical tokens in the $sub_{large}$ dataset. There is an increase in the proportion of numerical predictions around layer 9 in the post-MLP IPs, indicating the the model recognizes the subtraction task at the same time it recognizes the addition task. We find similar general trends in the $sub_{small}$ dataset. 
\paragraph{The model starts to converge on the result in mid- to late-layers.}
Figure \ref{fig:numericalproportion_sub} shows the average proportion of numerical tokens within the top 1 and top 10 IPs, on the $sub_{large}$ dataset. We find similar patterns to the addition task, where early- to mid-layers (layers 10 to 17) and mid- to late-layers (layers 29 to 43) reliably produce some numerical predictions. However, differently from the addition task, the model predicts a numerical token as the top prediction in earlier, starting from layer 36. 
Figure \ref{fig:absoluteErrorPredictions_sub} shows that, similar to the addition task, in the post-MLP IPs, the similarity between the predicted numerical tokens and the correct result is generally low in early and mid layers, but begins to increase incrementally after layer 34, corresponding to the layer where the correct result is assigned a higher probability (Figure \ref{fig:correct_result_sub}). There is a notable decrease in absolute error for the top 1 post-ATT predictions in layers 23 and 26, indicating that the model has a reasonable approximation of the magnitude of the correct result at that time.
\paragraph{The correct result is produced only in late layers.} 
The correct result typically emerges in the later layers of the model (Figure \ref{fig:correct_result_sub}). This is demonstrated by a significant increase in the correct token's probability and a corresponding decrease in its position, with the correct result appearing around layers 35-38 for large subtraction tasks. Similarly to the addition task, smaller subtraction tasks are solved earlier with the correct result emerging as early as layers 25-30.


\paragraph{Attention modules propagate information at specific layers.}
For the post-ATT IPs, we observe similar significant spikes in numerical predictions at specific layers as in the addition task (Figure \ref{fig:numericalProbmass_sub}).
\paragraph{Mostly, a single operand is attended to.}
Similarly to the addition task, we find that both operands are predicted as the highest probability (rank 1) token, in 55.2\% of queries. Specifically, operand 1 is the highest predicted token at the post-ATT representation of \textit{some} layer (mean layer 26.3) in 43\% of cases, and operand 2 (mean layer 19.6) in 32\%. This indicates that operands are in fact propagated to the final token position. We also find that in 66\% of cases, the propagation of operands is \textit{mutually exclusive}, such that when operand 1 is occurs as the top prediction in a post-ATT IP, operand 2 does not, and vice versa.
\section{Results GPT-2 XL}
After gaining a good understanding how GPT-Neox-20b solves simple arithmetic tasks, we now briefly investigate the internal mechanisms of GPT-2 XL \cite{radford2019languageGPT2xl} while generating predictions for the same tasks. 
As shown in Table \ref{tab:accuracy} GPT-2 XL is unable to generate the correct results for our datasets, so a comparison of the mechanisms between both models is of interest.

\begin{figure*}[t]
    \centering
    \begin{subfigure}[b]{0.475\textwidth}
        \centering
        \includegraphics[width=\textwidth]{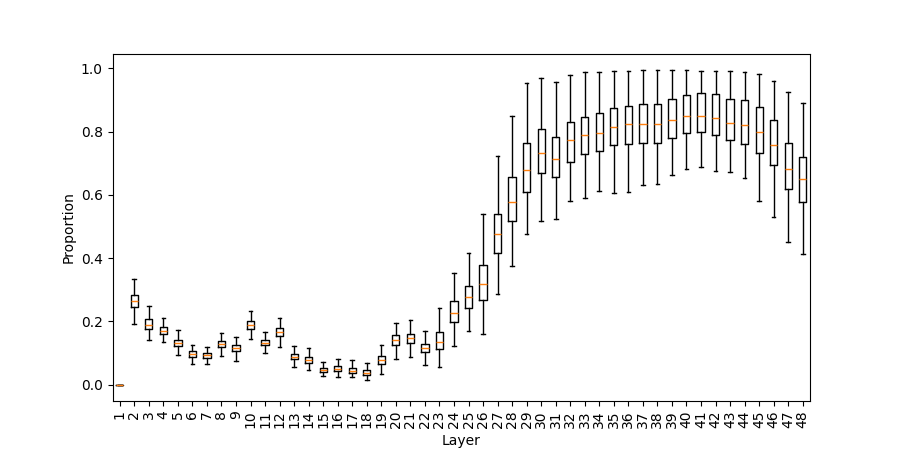}
        \caption[(a)]%
        {}    
    \end{subfigure}
    \hfill
    \begin{subfigure}[b]{0.475\textwidth}  
        \centering 
        \includegraphics[width=\textwidth]{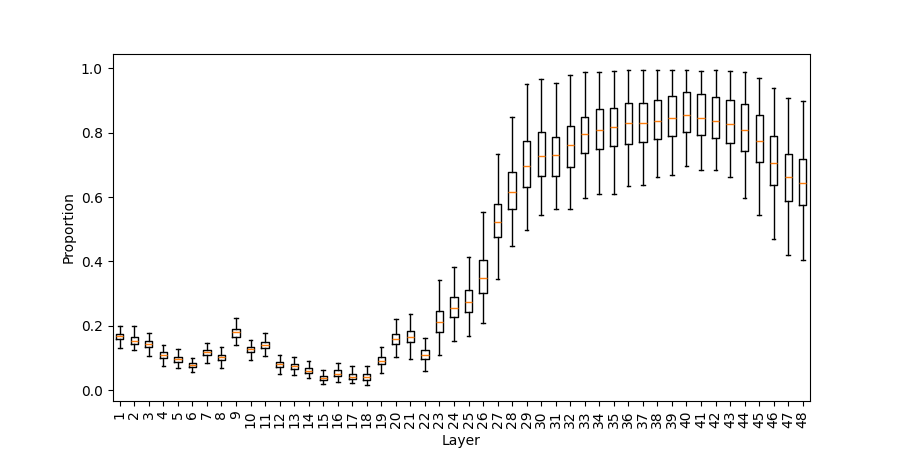}
        \caption[(b)]%
        {}   
    \end{subfigure}
    \caption[Proportion numerical addition]
    {GPT-2 XL: Combined probability mass of numerical tokens in the (a) post-ATT and (b) post-MLP intermediate predictions, averaged on the $add_{large}$ dataset.} 
    \label{fig:numericalProbmass_add_xl}
\end{figure*}

\begin{figure*}[t]
    \centering
    \begin{subfigure}[b]{0.475\textwidth}
        \centering
        \includegraphics[width=\textwidth]{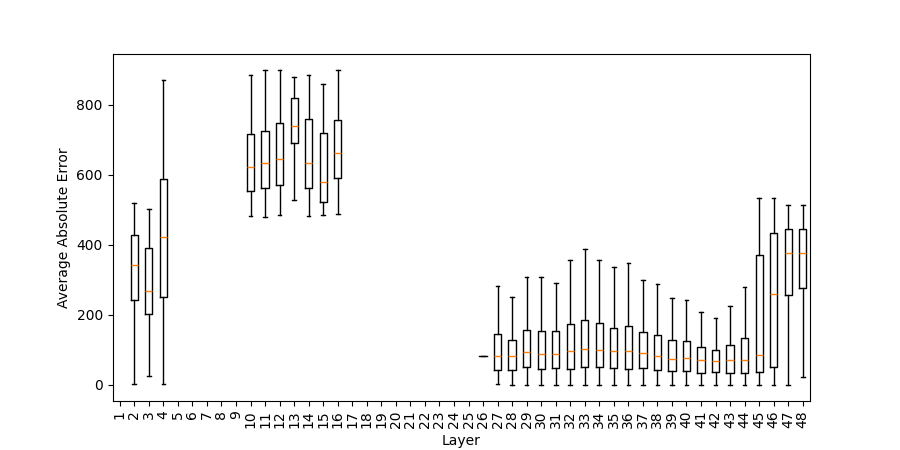}
        \caption[(a)]%
        {}   
    \end{subfigure}
    \hfill
    \begin{subfigure}[b]{0.475\textwidth}  
        \centering 
        \includegraphics[width=\textwidth]{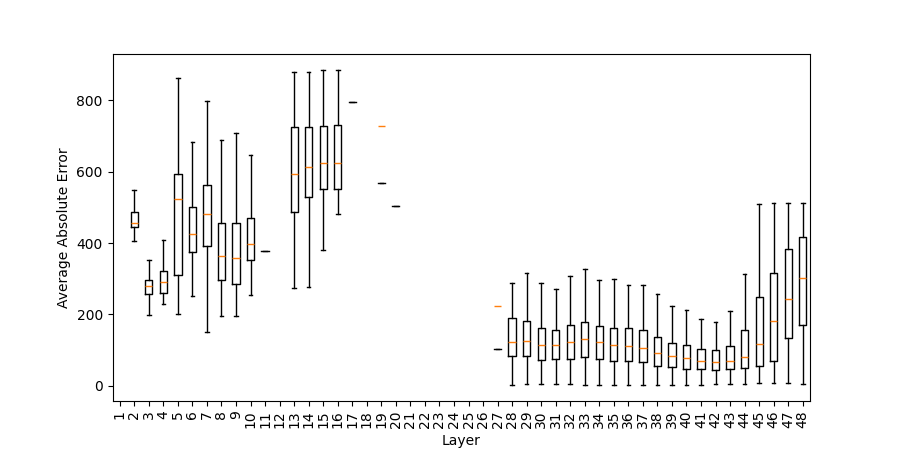}
        \caption[(b)]%
        {}  
    \end{subfigure}
    \vskip 0.5\baselineskip
    \begin{subfigure}[b]{0.475\textwidth}   
        \centering 
        \includegraphics[width=\textwidth]{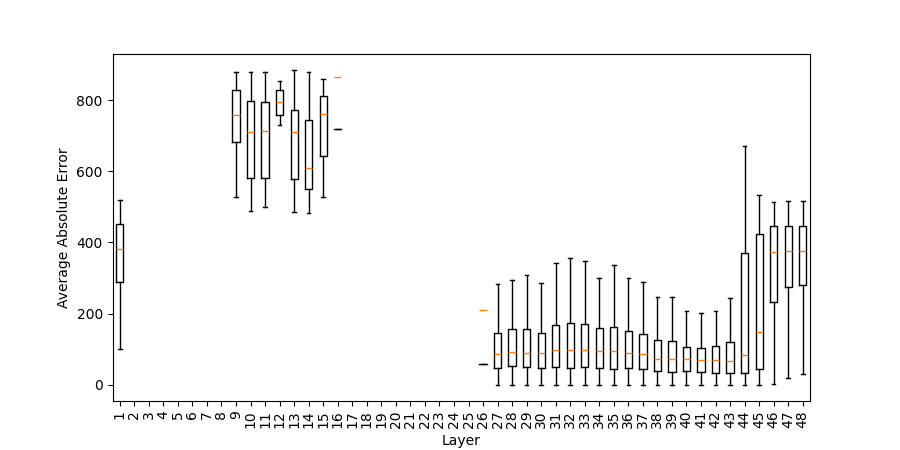}
        \caption[(c)]%
        {}   
    \end{subfigure}
    \hfill
    \begin{subfigure}[b]{0.475\textwidth}   
        \centering 
        \includegraphics[width=\textwidth]{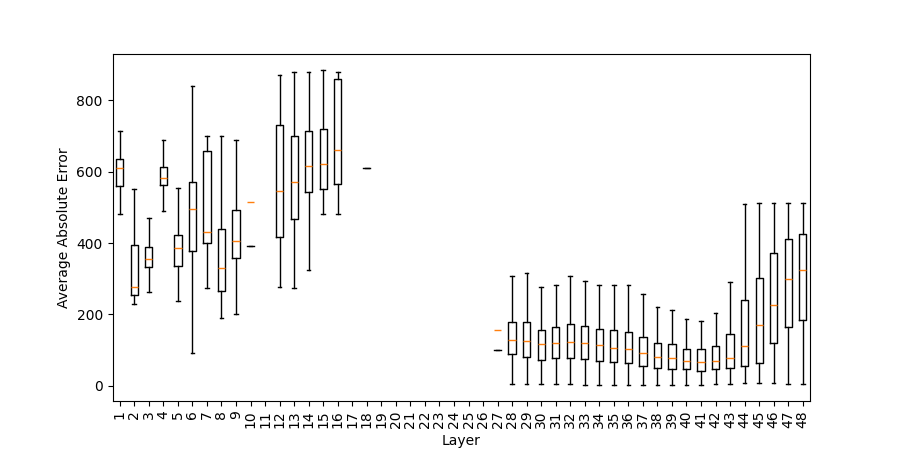}
        \caption[(d)]%
        {}  
    \end{subfigure}
    \caption[ absolute error numerical addition]
    {GPT-2 XL: Absolute error, i.e., difference to correct result, of numerical tokens in the (a) top 1 and (b) top 10 post-ATT intermediate predictions. And absolute error of numerical tokens in the (c) top 1 and (d) top 10 post-MLP intermediate predictions. All averaged on the $add_{large}$ dataset.} 
    \label{fig:absoluteErrorPredictions_add_xl}
\end{figure*}

\begin{figure*}[t]
    \centering
    \begin{subfigure}[b]{0.475\textwidth}
        \centering
        \includegraphics[width=\textwidth]{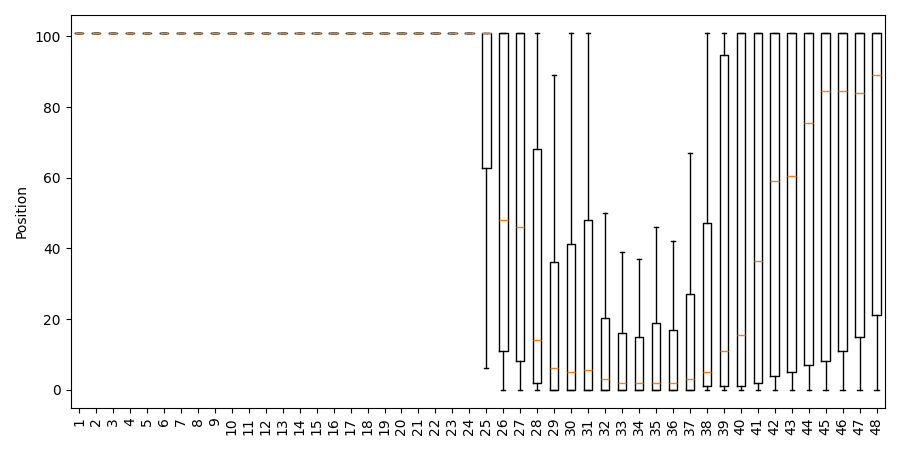}
        \caption[(a)]%
        {}    
    \end{subfigure}
    \hfill
    \begin{subfigure}[b]{0.475\textwidth}  
        \centering 
        \includegraphics[width=\textwidth]{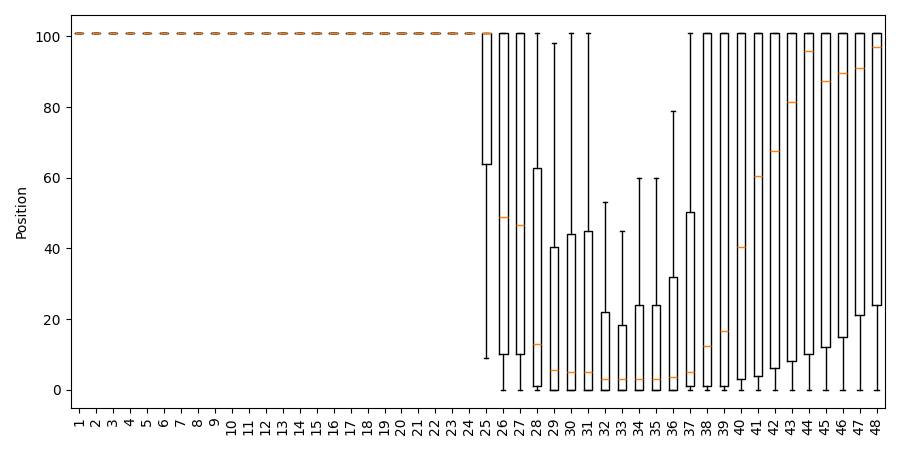}
        \caption[(b)]%
        {}    
    \end{subfigure}
    \caption[ Operands 'loaded']
    {GPT-2 XL: Position of operand 1 in the (a) post-ATT and (b) post-MLP prediction of intermediate layers, averaged over all data points in $add_{large}$.} 
    \label{fig:operands_xl}
\end{figure*}

\paragraph{The model predicts numerical tokens very early.} 
GPT-2 XL predicts numerical tokens early and frequently (Figure \ref{fig:numericalProbmass_add_xl}). It is however very unlikely that this signals a task understanding, as the model could not have deciphered the task to be solved in the very first layer, as no contextualization has been done. 
\paragraph{The model does not get close to the correct result.}
Figure \ref{fig:absoluteErrorPredictions_add_xl} shows that the top numerical prediction gets more similar in magnitude to the correct result in mid- to late-layers. However, the similarity decreases again in late layers, signaling that the model neither generates the correct result (Table \ref{tab:accuracy}) nor generate a result of correct magnitude.
\paragraph{The model shows comparable behavior post-MLP and post-ATT.}
Figure \ref{fig:numericalProbmass_add_xl} and Figure \ref{fig:absoluteErrorPredictions_add_xl} shows that the post-MLP and post-ATT predictions are very similar. This is a drastic difference to intermediate predictions generated by GPT-Neox-20b. We thus suspect that the division of tasks between attention and MLP modules that we find in GPT-Neox-20b is not present in GPT-2 XL. \\
\paragraph{Operands are propagated to the output token.}
We find that operands are propagated to the output token, as well as generated by MLP modules (Figure \ref{fig:operands_xl}). This suggests that attending to operands is not sufficient for the model to solve arithmetic tasks. We suspect that the lack of distinctive MLP and attention `behavior' hinders the generation of the correct result.

\end{document}